\documentclass[runningheads]{llncs}

\usepackage[T1]{fontenc}
\usepackage{graphicx}
\usepackage{makecell}
\usepackage[normalem]{ulem}
\usepackage{soul}
\usepackage{amsmath}

\usepackage{booktabs}
\usepackage{multirow}
 \usepackage{threeparttable}

\usepackage{xcolor}

 \usepackage{float} % 
\begin{document}
\title{A Comprehensive Survey on Deep Learning Solutions for 3D Flood Mapping}
\titlerunning{Deep Learning Based 3D Flood Mapping}

\author{Wenfeng Jia\inst{1}\orcidID{0000-0002-3996-5438} \and
Bin Liang\inst{2}\orcidID{0000-0002-6605-2167} \and
Yuxi Lu\inst{2}\orcidID{0009-0002-2220-7873} \and
Muhammad Arif Khan\inst{1}\orcidID{0000-0001-6112-8874} \and
Lihong Zheng\thanks{Corresponding author: lzheng@csu.edu.au}\inst{1}\orcidID{0000-0001-5728-4356}}
\authorrunning{W. Jia et al.}
\institute{Charles Sturt University, New South Wales, Australia  \and
University of Technology Sydney, New South Wales, Australia}

\maketitle              % typeset the header of the contribution

\begin{abstract}
Flooding remains a major global challenge, worsened by climate change and urbanization, demanding advanced solutions for effective disaster management. While traditional 2D flood mapping techniques provide limited insights, 3D flood mapping, powered by deep learning (DL), offers enhanced capabilities by integrating flood extent and depth. 
This paper presents a comprehensive survey of deep learning-based 3D flood mapping, emphasizing its advancements over 2D maps by integrating flood extent and depth for effective disaster management and urban planning. The survey categorizes deep learning techniques into task decomposition and end-to-end approaches, applicable to both static and dynamic flood features. We compare key DL architectures, highlighting their respective roles in enhancing prediction accuracy and computational efficiency. Additionally, this work explores diverse data sources such as digital elevation models, satellite imagery, rainfall, and simulated data, outlining their roles in 3D flood mapping. The applications reviewed range from real-time flood prediction to long-term urban planning and risk assessment. However, significant challenges persist, including data scarcity, model interpretability, and integration with traditional hydrodynamic models. This survey concludes by suggesting future directions to address these limitations, focusing on enhanced datasets, improved models, and policy implications for flood management. This survey aims to guide researchers and practitioners in leveraging DL techniques for more robust and reliable 3D flood mapping, fostering improved flood management strategies.

\end{abstract}
\section{Introduction}
%\subsection{Background and Motivation}
Flooding has been a persistent threat to human society since its inception, and its impact has not diminished in modern times. Research indicates that flood frequency has shown an upward trend over the past 30 years~\cite{Flood2022increase}. Meanwhile, global urbanization has increased the concentration of populations and assets in flood-prone areas, leading to more severe damage when floods occur. According to the global hazards database, between 2018 and 2022, flood disasters caused more than 20,000 deaths or missing people, with direct economic losses exceeding \$189.8 billion USD~\cite{Li2024Flood_DL_review}. Currently, 1.81 billion people (approximately 23\% of the global population) are directly exposed to flood risks~\cite{Flood2022affect}. 

Given the significant damage caused by floods, reliable flood mapping techniques are critical for disaster management and risk assessment. They offer the potential to significantly reduce the threat that floods pose to life and property. 
Traditional 2D flood mapping, which focuses solely on flood extent, is inadequate for capturing the full scope of flood risks. The inclusion of flood depth through 3D mapping provides critical insights into flood severity, enabling accurate risk assessments and targeted disaster management.
Before deep learning (DL), physical models were the main tools for flood mapping, offering detailed simulations but requiring complex setups, high-quality data, and substantial computational resources~\cite{Flood_hazard_review_traditional_2021}.
With rapid advancements in computer technology, machine learning has greatly improved. To overcome the low computational efficiency of traditional physical models, some researchers have turned to physics-guided machine learning for flood mapping~\cite{Karim2023HydrodynamicML_review}.
In recent years, DL, with its ability to automatically extract features and handle complex datasets~\cite{chen2020spatial,han2023survey,kussul2017deep}, has emerged as a faster and more efficient alternative for 3D flood mapping~\cite{Seleem2023CNN_RF_Compare}. 

Given the current lack of review articles on 3D flood mapping using DL techniques, this survey presents a comprehensive analysis of recent advancements in the field.
By categorizing DL techniques into \textbf{task decomposition} and \textbf{end-to-end} methods, this paper explores their applicability to \textbf{static} and \textbf{dynamic} flood mapping. The review also explores the challenges of data scarcity, model generalizability, and practical implementation, offering insights into future directions for this rapidly evolving field.
% Literature Retrieval Criteria 
To identify high-quality literature and gain a comprehensive understanding of the research methods and current status of 3D flood mapping, this survey conducted a systematic search using WebofScience and Google Scholar. The search criteria were as follows: 
\begin{itemize}
    \item WebofScience: (TS=(flood mapping) AND TS=(deep learning) AND 
    (TS=(3D) OR TS=(flood depth))) AND PY=(2020-2025)
    \item Google Scholar: ``flood mapping'' ``deep learning''  (OR 3D  OR ``flood depth'')  -review -survey. (Since 2020)
\end{itemize}
The initial search retrieved 55 articles from WebofScience and 58 from Google Scholar. After removing duplicates and weakly relevant papers, 39 articles were ultimately selected for analysis in this survey.

Based on this curated collection, we outline the structure of the survey as follows: Section 2 covers the foundations of 3D flood mapping, including problem statements, method comparisons, and an introduction to data sources. Section 3 examines DL models categorized by their methodologies and the flood features they address. Section 4 highlights application directions, while Section 5 reviews model metrics. Section 6 discusses challenges and outlines potential future research directions. Finally, Section 7 concludes the survey.

\section{The Foundations of DL-based 3D Flood Mapping}
This section provides a foundational understanding of 3D flood mapping, covering the problem statement, a comparison of existing methods, and the primary data sources used in DL-based 3D flood mapping.

\subsection{Problem Statement}
\begin{figure}
    \centering
    \includegraphics[width=0.86\linewidth]
        {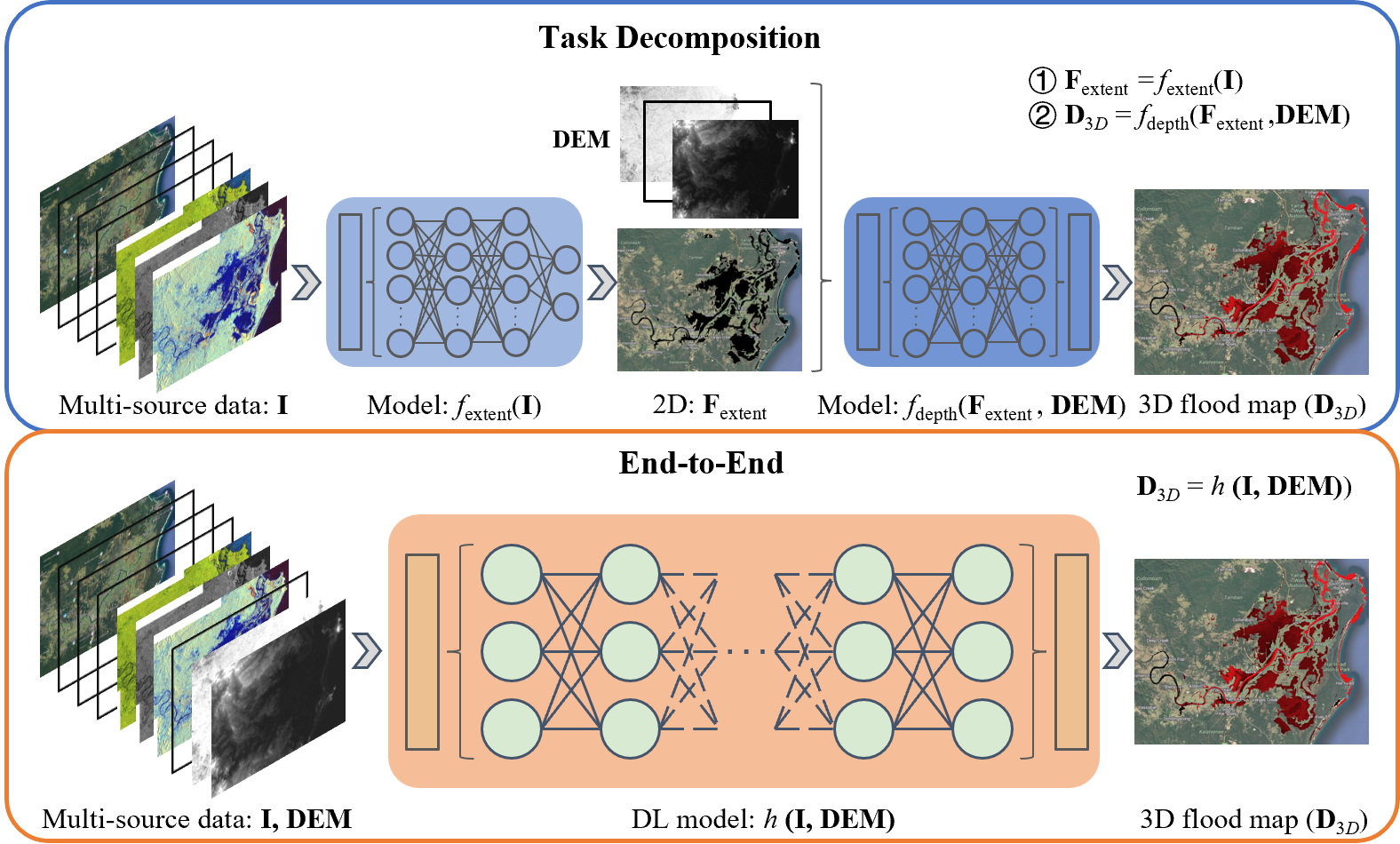}
    \caption{Task decomposition and end-to-end methods for 3D flood mapping.}
    \label{fig:methods}
\end{figure}

Flood mapping is essential for disaster management, aiming to predict flood extent and depth accurately to produce a 3D flood map. As shown in Fig.~\ref{fig:methods}, two DL-based methodologies, task decomposition methods, and end-to-end methods offer distinct approaches to solving this problem. This section formalizes these methodologies with consistent notation and generalized loss functions.

In task decomposition methods, the flood mapping process is divided into two steps. First, the flood extent detection step predicts the 2D flooded area. Second, the flood depth estimation step calculates the depth of flooding in the detected area and produces the final 3D flood map:

\begin{equation}
\mathbf{D}_{3D} = f_{\text{depth}}( \mathbf{F}_{\text{extent}}, \mathbf{DEM})
\label{D3D}
\end{equation}
where: 
\begin{itemize}
    \item \( f_{\text{depth}} \) is a flood depth estimation model predicting depth and generating 3D flood maps using $\mathbf{F}_{\text{extent}}$ and digital elevation model ($\mathbf{DEM}$).
    
    \item $\mathbf{F_{\text{extent}}}$ is a flood extent detection model predicting the 2D maps. It is defined as $\mathbf{F_{\text{extent}}}=f_{\text{extent}}(\mathbf{I})$, and $\mathbf{I}$ represents the input data, such as satellite, unmanned aerial vehicle (UAV), and rainfall data.
\end{itemize}

For the two separate steps, flood extent detection and flood depth estimation, each uses its independent loss function for the corresponding model training:
\begin{equation}
\begin{aligned}
\mathcal{L}_{\text{extent}} &= \mathcal{L}_{\text{classification}} \left( \mathbf{F}_{\text{extent}}, \mathbf{F}_{\text{extent}}^{\text{true}} \right) \\
\mathcal{L}_{\text{depth}} &= \mathcal{L}_{\text{regression}} \left( \mathbf{D}_{\text{depth}}, \mathbf{D}_{\text{depth}}^{\text{true}} \right)
\end{aligned}
\label{task_loss_functions}
\end{equation}

As shown in the Equation~\ref{task_loss_functions}, $\mathcal{L}_{\text{extent}}$ is a general classification loss function, which measures how well the model predicts the flooded area. Common loss functions include Binary Cross-Entropy, Dice Loss, or Focal Loss.
$\mathcal{L}_{\text{depth}}$ is a general regression loss function, which evaluates the accuracy of the predicted depth. Common options include Mean Squared Error and Mean Absolute Error.

In the end-to-end methods, a single unified model predicts the full 3D map:
\begin{equation}
    \mathbf{D}_{3D} = h(\mathbf{I}, \mathbf{DEM})
    \label{D3D_end}
\end{equation}
where:
\begin{itemize}
    \item $\mathbf{I}$ is the input data (e.g., satellite, UAV or rainfall data).
    \item $\mathbf{DEM}$ is the digital elevation model.
    \item \( h \) is the end-to-end model that predicts the full 3D flood map.
\end{itemize}

The loss function for the end-to-end methods combines components for flood extent and depth prediction:
\begin{equation}
\mathcal{L}_{\text{e2e}} = \lambda \mathcal{L}_{\text{classification}} \left( \mathbf{D}_{\text{extent}}, \mathbf{F}_{\text{extent}}^{\text{true}} \right)
+ \gamma \mathcal{L}_{\text{regression}} \left( \mathbf{D}_{\text{depth}}, \mathbf{D}_{\text{depth}}^{\text{true}} \right)
\end{equation}
where:
\begin{itemize}
    \item $\mathcal{L}_{\text{classification}}$ and $\mathcal{L}_{\text{regression}}$ are general loss functions.
    \item $\lambda$ and $\gamma$ are weights balancing the classification and regression components.
\end{itemize}

The end-to-end methods simplify the process by using a single model to predict the entire 3D flood map, eliminating the need for intermediate steps.

\subsection{Comparison of DL-based 3D Flood Mapping Methods}
Building on the problem statement, this subsection compares existing methodologies to highlight their strengths and limitations. Task decomposition methods break 3D flood mapping into two separate subtasks. In contrast, end-to-end methods use one model to learn flood extent and depth directly from multi-source data.
Table~\ref{tab:comparison} provides a comparison of the two methods.

\begin{table}
\centering
\caption{Comparison of task decomposition and end-to-end methods.}
\label{tab:comparison}
\begin{tabular}{@{}>{\bfseries}p{2.4cm}p{5.2cm}p{4.6cm}@{}}
\toprule
Aspect               & \textbf{Task Decomposition Methods}                       & \textbf{End-to-End Methods}                       \\  \hline \addlinespace
Workflow             & Two stages: $\mathbf{F}_{\text{extent}}$ and $\mathbf{D}_{\text{depth}}$. & Single unified model $h(\mathbf{I}, \mathbf{DEM})$.          \\ \addlinespace
Loss Functions       & Separate: $\mathcal{L}_{\text{classification}}$ and $\mathcal{L}_{\text{regression}}$. & Combined: $\mathcal{L}_{\text{e2e}}$.            \\ \addlinespace
Interpretability     & Intermediate outputs available.             & Direct output, less interpretable.     \\ \addlinespace
Flexibility          & Modular and adaptable.                           & Requires re-training for changes.      \\ \addlinespace
Performance          & Errors can spread between subtasks.             & Joint optimization often improves accuracy.\\ \addlinespace
Efficiency           & Sequential steps slow down inference.               & One model speeds up inference.            \\ \bottomrule
\end{tabular}
\end{table}

As shown in Table~\ref{tab:comparison}, task decomposition methods allow for flexible optimization of each subtask and make it easy to integrate with current 2D flood mapping research. However, since these subtasks are performed in sequence, error propagation between them can reduce the accuracy of the final results. Additionally, the sequential nature increases computation time.
In contrast, the end-to-end methods use a single model to directly learn patterns of flood extent and depth from multi-source data. This approach eliminates intermediate steps, avoiding human-induced errors and error propagation while improving computational efficiency. However, the lack of intermediate outputs reduces the model's flexibility, and re-training may be required for scenario changes.

\subsection{Data Sources for 3D Flood Mapping}
This subsection reviews the data sources that support these methods and presents the resolution, coverage, and information provided by different data types.

Common 3D data sources include satellites, UAVs, Light Detection and Ranging (LiDAR), DEM, rainfall, and water gauges. These data can be processed through DL and GIS technologies to create accurate flood models. 
Some typical data sources are shown in Table \ref{dataSource}.

\begin{table}[h]
\begin{threeparttable} 
    \caption{Summary of key data sources}
    \label{dataSource}
\begin{tabular}{@{}>{\bfseries}lllll@{}}
\toprule
\ Sources & \textbf{Spatial Res} & \textbf{Temporal Res} & \multicolumn{1}{c}{\textbf{Coverage}} & \multicolumn{1}{c}{\textbf{Notes}} \\ \midrule
Satellites & 5--100 m & 5--16 days & Global/Regional & Multispectral/Hyperspectral. \\
UAVs & <0.1 m& On-demand & Local& RGB, Multispectral. \\
DEM & 1--30 m& >1 year & Global/Regional & Terrain data. \\
LiDAR & <1 m& On-demand & Local & Terrain data (higher cost). \\
Rainfall & 1k--10km & Minutes/Hours & Global/Regional & Rain gauges or simulation. \\
Gauge & Point scale & - & Point level& Monitoring of water level \\ \bottomrule

\end{tabular}

\end{threeparttable}
\end{table}

%\subsection{Discussion of data}
Certain data sources, such as satellite and UAV data, provide 2D information, while others, like DEMs, are crucial for estimating flood depth. Rainfall records contribute temporal features, and water gauge measurements can offer ground truth for flooding in specific regions.

However, these data sources differ not only in format but also in spatial coverage, resolution, and other characteristics. While the complementarity of different data sources offers multifaceted support for 3D flood mapping, it also introduces challenges in multi-data fusion.

\section{DL Models for 3D Flood mapping}
This section focuses on reviewing DL models for 3D flood mapping and classifies these approaches into two types: task decomposition and end-to-end methods, both applicable to static and dynamic scenarios.

\begin{figure}
    \centering
    \includegraphics[width=0.8\linewidth]
        {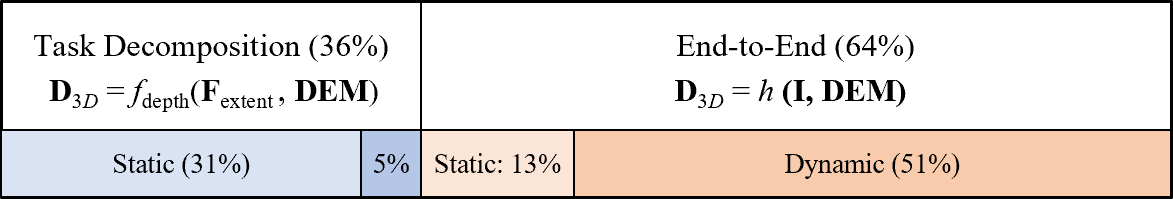}
    \caption{The proportion of the two methods in DL-based 3D flood mapping.}
    \label{fig:method}
\end{figure}

% \paragraph{Static vs. Dynamic Flood Mapping}
Static flood mapping focuses on predicting floods at a specific time, emphasizing spatial characteristics. In contrast, dynamic flood mapping captures the temporal features of floods, integrating both spatial and temporal dimensions. The proportion of two methods in DL-based 3D flood mapping is shown in Fig.~\ref{fig:method}.

As shown in Fig.~\ref{fig:method}, among the classifications, studies based on end-to-end methods for dynamic scenarios are the most common, accounting for 51\%, while dynamic mapping based on task decomposition is the least common, representing only 5\%.

\subsection{Task Decomposition Methods}
According to Fig.~\ref{fig:method}, task decomposition can be categorized into Static Flood Mapping and Dynamic Flood Mapping based on the type of output mapping.

\paragraph{Static Flood Mapping}
%  D_extent
In static flood mapping, various models can handle the $\mathbf{F}_{\text{extent}}$ subtask. Convolutional Neural Networks (CNNs) are popular because they automatically extract spatial features from large flood datasets.For example, Gebrehiwot et al.~\cite{Gebrehiwot2021floodDepth_UAV_DEM_FCN} utilized a Fully Convolutional Neural Network (FCN-8s) to extract $\mathbf{F}_{\text{extent}}$ and extended this to $\mathbf{D}_{\text{3D}}$  by incorporating DEM. They compared various methods, including Structure from Motion~\cite{Gebrehiwot2020floodDepth_UAV_DEM_VGG16}, and the Geomorphic Flood Index~\cite{Gebrehiwot2022floodDepth_UAV_DEM_FCN}, concluding that integrating DL models with DEM and UAV imagery significantly enhances 3D flood mapping accuracy.

Building on this method, Tan et al.~\cite{Tan2024upscalingDEM_CNN} employed the DeepLabV3 model with transfer learning for  $\mathbf{F}_{\text{extent}}$ and used ResNet-34 to upscale low-resolution DEM data. This approach addressed the challenge of limited high-resolution data, enabling the generation of high-resolution $\mathbf{D}_{\text{3D}}$.

% D_depth
DL models can be used not only for detecting $\mathbf{F}_{\text{extent}}$ and enhancing flood data quality, but also for estimating $\mathbf{D}_{\text{depth}}$. Tokoya et al.~\cite{Yokoya2022DL_modelFusion} proposed a framework combining Attention U-Net and LinkNet to estimate flood depth and terrain deformation. Because of the powerful pattern learning capabilities of DL models, this approach can even compensate for omission or misclassification parts in remote sensing methods.

% D_extent& D_depth
Additionally, DL models can be used for both $\mathbf{F_{\text{extent}}}$ detection and $\mathbf{D}_{\text{depth}}$ estimation.
Generative Adversarial Networks (GANs) provide structural flexibility for complex scenarios in  $\mathbf{D}_{\text{3D}}$. Based on this characteristic, do Lago et al.~\cite{do2023CGAN} proposed a dual-generator GAN model adhering to flood volume conservation principles. One generator identified  $\mathbf{F}_{\text{extent}}$, while the other estimated  $\mathbf{D}_{\text{depth}}$. This approach achieved simulation speeds 250 times faster than the hydrodynamic model HEC-RAS~\cite{do2023CGAN}.

% others
In addition to the above, DL models can also have other applications in task decomposition methods. For example, they can assist in data preprocessing to support hydrodynamic models. Fang et al.~\cite{Fang2024DL_MIKE} developed a workflow using the RandLA-Net to classify UAV data, automatically assigning runoff parameters and grid resolutions to different areas. These classifications were subsequently utilized by the hydrodynamic model MIKE 21 to simulate floods. This framework reduces manual configuration efforts and improves simulation accuracy.

Although various DL models can be used, static mapping typically relies on spatial data alone, without temporal information. In contrast, the dynamic features of floods are often closely tied to temporal factors like rainfall~\cite{Xie2021Hybrid-ANN_data-sparse}, which can provide a better understanding of flood events.

\paragraph{Dynamic Flood Mapping}
% \paragraph{Long Short-Term Memory (LSTM)}
Recurrent Neural Networks (RNNs) are DL models designed for sequential data. However, traditional RNNs suffer from vanishing and exploding gradient problems. To address these challenges, the Long Short-Term Memory (LSTM) was proposed, which can learn long-term dependencies~\cite{Sherstinsky2020RNN_LSTM}. Zhou Y. et al.~\cite{Zhou2021DL_SRR_TUFLOW_LSTM} introduced a Spatial Reduction and Reconstruction (SRR) framework combining LSTM, greatly enhancing computational efficiency by selecting and reconstructing key feature points. The SRR-DL framework achieved speeds 21 times faster than the hydrodynamic model (TUFLOW). 

% \paragraph{CNNs for Temporal Features}
Though traditionally spatial, CNNs can incorporate temporal information by processing serialized temporal rainfall data. Zhou Y. et al.~\cite{Zhou2022USRR1DCNN} enhanced their SRR framework~\cite{Zhou2021DL_SRR_TUFLOW_LSTM} by using U-Net, achieving higher accuracy and computational efficiency (98 times faster than TUFLOW).

\subsection{End-to-End Methods}
While task decomposition methods could integrate with current 2D flood mapping research, the lack of ground truth for flood depth presents significant challenges in DL training and 3D evaluation. Therefore, many researchers prefer to generate large amounts of flood simulation data using hydrodynamic models, which enable the use of end-to-end approaches for DL-based 3D flood mapping, as described in Equation~\ref{D3D_end}.

\paragraph{Static flood mapping}
% \paragraph{CNN:}
CNNs could be used in end-to-end methods for static flood mapping.
Hosseiny et al.~\cite{Hosseiny2021flooddepth_UNet} modified U-Net to predict $\mathbf{D}_{\text{3D}}$ flood discharge data, using hydraulic model (iRIC) simulations as ground truth. This approach reduced manual intervention, and achied a 29\% improvement in flood depth prediction accuracy compared to task decomposition and MLP-based methods~\cite{Hosseiny2020flooddepth_MLP-RF}.

Guo et al.~\cite{Guo2021CNN_CADDIES} treated flood depth prediction as an image transformation task, integrating spatial and temporal inputs to generate maximum flood depth predictions. This approach reduced computational time to 0.5\% of that required by the cellular-automata flood model (CADDIES). The authors~\cite{Guo2022catchment} further explored how catchment areas affect CNN-based flood mapping, finding that larger receptive fields improve accuracy.

\paragraph{Dynamic Flood Mapping}
% \paragraph{LSTM:}
Building on LSTM models, researchers have demonstrated that DL methods are significantly more computationally efficient than hydraulic models in dynamic flood mapping. Zhou, Q et al.~\cite{Zhou2023LSTMRapidMIKE21} combined rainfall time series data, DEM, LSTM, and Bayesian optimization to develop a framework for rapid $\mathbf{D}_{3D}$. This framework operates 19,585 times faster than the hydraulic model (MIKE 21) with a relative error of 9.5\%.

% \paragraph{CNN:}
Many other studies have shown that DL models offer high computational efficiency than hydraulic models. For instance, Kabir, S. et al.~\cite{Kabir2020CNN_LISFLOOD-FP_Spatial_Tempporal} utilized simulation data from a hydraulic model (LISFLOOD-FP) to estimate $\mathbf{D}_{3D}$ using a basic 1D CNN model. This model effectively simulates floods while significantly reducing computational costs. Several researchers have conducted similar studies~\cite{EL2024CNNRealTime,Guo2021CNN_CADDIES,Löwe2021UFlood}, highlighting the efficiency of DL models.

To further enhance the extraction of dynamic flood features, some researchers have integrated attention mechanisms into CNN models. Shao Y et al.~\cite{Shao2024CRU} introduced CRU-Net, a U-Net that integrates Convolutional Block Attention Module and Residual Block to predict $\mathbf{D}_{3D}$ in urban areas. Similarly, Chaudhery et al.\cite{Chaudhary2024AttentionUTAE} combined U-Net with temporal attention mechanisms, showcasing both the accuracy and computational efficiency of DL models. 

In 3D flood mapping, the importance of DEM cannot be overstated due to its provision of crucial elevation information. To investigate the impact of DEM on $\mathbf{D}_{3D}$, Fereshtehpour et al.~\cite{Fereshtehpour2024CNN_DEM_DTM} investigated the influence of DEM types and resolutions using a 1D-CNN model. Their findings revealed that digital terrain models outperform digital surface models in accuracy. Additionally, while low-resolution DEMs reduce the precision of DL models, they remain highly effective for applications requiring rapid 3D flood mapping.

% \paragraph{Graph Neural Networks (GNNs)}
Graph Neural Networks (GNNs) represent data as graphs, where nodes represent entities like geographical locations and edges show their relationships. By propagating information across nodes, GNNs update node features and learn the overall graph structure. Bentivoglio et al.~\cite{Bentivoglio2023SWE-GNN} used this approach to develop the Shallow Water Equation–Graph Neural Network (SWE–GNN) model for predicting water depth and discharges. Its propagation rule, based on shallow water equations, enhances the framework's interpretability.

\subsection{Discussions}
DL-based 3D flood mapping offers advantages over traditional methods, including better computational efficiency and scalability. Task decomposition methods are often used for static mapping and typically rely on measured data. These methods are modular, making it easier to integrate with other technologies. In contrast, end-to-end approaches are suited for dynamic mapping and primarily use simulated data. They simplify the process by employing a single unified model to generate 3D flood maps. More researchers are adopting end-to-end methods and simulated data for 3D flood mapping studies. However, their reliance on simulated data can limit their application in the real world.

Meanwhile, DL models demonstrate versatility across various stages of the 3D flood mapping workflow. CNNs are effective for capturing spatial and temporal features, while LSTMs are well-suited for modeling temporal dynamics. Advanced architectures, such as GANs and GNNs, further enhance the ability to handle complex scenarios. Moreover, DL models are not only used for flood extent and depth estimation but also play a critical role in data augmentation and preprocessing, highlighting their adaptability to diverse tasks and datasets.

\section{Applications of 3D Flood Mapping with DL}
\begin{figure} [H]
    \centering
    \includegraphics[width=0.9\linewidth]{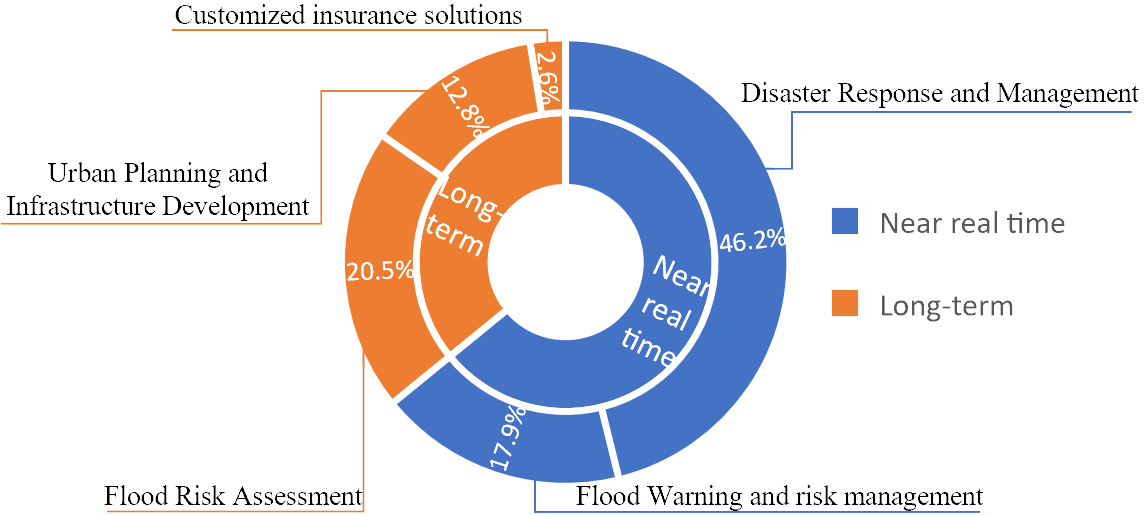}
    \caption{The proportion of different types of applications in 3D flood mapping.}
    \label{fig:Applications}
\end{figure}

This section highlights the applications of DL-based 3D flood mapping. The 3D flood maps provide not only flood extent but also key details on depth and hazard levels, enabling comprehensive risk assessments.
These capabilities support near real-time disaster response, long-term flood prediction, and urban planning. Near real-time applications address flood events as they occur or shortly after, while long-term applications cover scenarios over years or decades.
Fig.~\ref{fig:Applications} summarizes the applications of DL-based 3D flood mapping. 

As shown in Fig.~\ref{fig:Applications}, most studies focus on near real-time applications, especially disaster response and flood management, which make up over 46\% of the cases. This underscores the need for rapid 3D flood mapping, a key strength of DL models. In contrast, customized insurance solutions account for less than 3\% of the studies. Although this approach is a passive response to flooding and is underexplored, it has potential to provide fair protection for residents and assets. 

\subsection{Near Real-Time Flood Prediction}
The computational efficiency of DL models enables the generation of high-resolution 3D flood maps shortly after flood events~\cite{dong2021hybrid_application_near_real,kao2021fusing_application_near_real}. However, due to the delays from data platforms, such mapping is typically classified as near real-time. To overcome this limitation, some studies have explored rainfall-driven real-time flood prediction as a solution~\cite{chang2020artificial_application_near_real,ravuri2021skilful_application_near_real}.

% \subsubsection{Flood Warning and risk management}
By combining DL models with rainfall prediction, it is possible to perform 3D flood maps in advance for flood warnings~\cite{chang2020artificial_application_near_real,Chaudhary2022FloodEventManagement,Pianforini2024RealTime,Zhang2023RealTime}. For instance, Chaudhary et al.~\cite{Chaudhary2024AttentionUTAE}, Bentivoglio et al.~\cite{Bentivoglio2023SWE-GNN}, and El Baida et al.~\cite{EL2024CNNRealTime} can generate high-precision 3D flood maps 2 to several tens of hours in advance based on rainfall data. Combining predicted 3D flood maps for different regions with risk data, such as population density, decision-makers can make decisions more accurately.

% \subsubsection{Disaster Response and Management}
Detailed 3D flood information is essential for disaster response, enabling targeted actions during flood disasters~\cite{He2023RiskManagement}. Tan et al.~\cite{Tan2024upscalingDEM_CNN} demonstrated in Calgary, Canada, that DL models can generate high-resolution 3D flood maps even with low-resolution data. Furthermore, Li et al.~\cite{Li2022FloodWebsite} utilized GIS methods to develop an online portal for the automated processing and visualization of global 3D flood maps. This allows decision-makers to prioritize severely impacted regions and facilitate efficient rescue operations for people and property in danger zones.

\subsection{Long Term Flood Prediction}
Additionally, this survey explores the use of 3D flood mapping in strategic planning and long-term flood risk assessment.
DL's ability to process large datasets makes it well-suited for 3D flood susceptibility maps based on historical flood data. These maps enable authorities to assess flood risk, customize insurance policies, and proactively plan urban flood control infrastructure, minimizing future economic losses and casualties.

% \subsubsection{Flood Risk Assessment}
Long-term historical data such as hydrology and flooding records can be used for flood risk assessment. Based on the 3D risk assessment, measures can be taken to reduce flood hazards~\cite{Cosrache2020FlashFloodSusceptibility,Yu2024RiskAssessment}. Bui et al.~\cite{bui2020novel} proposed a DL approach for flash flood risk assessment in Northwestern Vietnam. Their model provided valuable tools for land-use management, offering insights to reduce flood impacts.
 
% \subsubsection{Customized insurance solutions}
Since 3D flood susceptibility maps provide richer information than 2D maps, customized insurance solutions for severely affected flood areas become possible. Chen et al.~\cite{chen2021towards_application_near_real} studied rainfall, DEM, and historical flood events  to develop 3D flood risk maps. Their study identified urban coastal areas as high-risk zones and emphasized the importance of adaptive measures, such as raising levee heights and improving insurance systems, to cope with increasing flood threats.

% \subsubsection{Urban Planning and Infrastructure Development}
 With the help of 3D flood susceptibility maps, targeted urban planning and infrastructure development for floods have seen increasing applications~\cite{Wu2020UrbanFloodControl}. Panahi et al.\cite{panahi2021deep_application_long-term} created flood probability maps for Golestan province in Iran, revealing that nearly 40\% of the area is flood-prone and guiding planning efforts. Similarly, Khosravi et al.\cite{khosravi2020convolutional_application_long-term} mapped flood susceptibility for Iran, showing that 12\% of the country is at risk. These maps provide essential data for flood control planning, helping high-risk regions mitigate future impacts.

\section{Evaluation Metrics and Model Performance}
This section details the criteria used to assess DL-based 3D mapping methods and provides a comparative analysis of evaluation metrics from the literature.
\subsection{Evaluation Metrics}
% \subsection{Accuracy and Precision}
Accurately evaluating model performance is crucial, regardless of the type of research methods used. Compared to 2D flood maps, 3D maps include depth information, which requires additional factors to accurately assess both accuracy and precision. Fortunately, the geometric extent and elevation information of floods in 3D models can be decoupled and evaluated separately.

% \subsubsection{2D Evaluation}
The evaluation of the 2D extent of floods can be treated as a typical binary classification task. There are various evaluation metrics, such as the Recall~\cite{Kabir2020CNN_LISFLOOD-FP_Spatial_Tempporal}, and the Intersection over Union (IoU). Since the classification of flooded regions often demands high spatial accuracy, the IoU metric is commonly used to evaluate the spatial accuracy of flood predictions~\cite{Tan2024upscalingDEM_CNN,Yokoya2022DL_modelFusion}.
% \subsubsection{Elevation Evaluation}
In 3D flood mapping, the model's prediction of flood depth is considered a regression task. The overall assessment of flood elevation results can be conducted using regression error metrics such as Root Mean Square Error (RMSE)~\cite{EL2024CNNRealTime,Fereshtehpour2024CNN_DEM_DTM,Zhou2023LSTMRapidMIKE21} and Mean Absolute Error (MAE)~\cite{Bentivoglio2023SWE-GNN,Chaudhary2024AttentionUTAE}.

% \subsection{Computational Efficiency}
In addition to examining the performance of 3D flood mapping models, it is essential to consider their computational efficiency, as this is vital for large-scale and near-real-time flood prediction. Since many researchers use  hydrodynamic models to provide ground truth or as a benchmark for comparison, DL methods are often evaluated against these models in terms of simulation time to assess their computational efficiency quantitatively~\cite{do2023CGAN,Zhou2022USRR1DCNN}. 

\subsection{Model Performance of DL based 3D Flood Mapping}
Table \ref{summary} summarizes the evaluation metrics for DL-based 3D flood mapping and includes details on the methods, data, models, and more. Due to space constraints, only a subset of the literature is presented.

\begin{table}
\begin{threeparttable} 
\scriptsize
    \caption{Summary of the evaluation metrics for DL based 3D flood mapping}
    \label{summary}
\begin{tabular}{@{}cclllllllll@{}}
\toprule

\multirow{3}{*}{Method} & \multirow{3}{*}{Feature} & \multicolumn{1}{c|}{\multirow{3}{*}{Paper}} & \multicolumn{2}{c|}{Data} & \multicolumn{1}{c|}{\multirow{3}{*}{Models}} & \multicolumn{4}{c|}{Metrics} & \multicolumn{1}{c}{\multirow{3}{*}{\begin{tabular}[c]{@{}c@{}}Time\\ Efficiency\footnotemark[2]\end{tabular}}} \\ \cmidrule(lr){4-5} \cmidrule(lr){7-10}
 &  & \multicolumn{1}{c|}{} & \multicolumn{1}{c|}{\multirow{2}{*}{Sources}} & \multicolumn{1}{c|}{\multirow{2}{*}{\begin{tabular}[c]{@{}c@{}}Res\footnotemark[1]\\ (m)\end{tabular}}} & \multicolumn{1}{c|}{} & \multicolumn{2}{c|}{Extent(\%)} & \multicolumn{2}{c|}{Depth (cm)} & \multicolumn{1}{c}{} \\ \cmidrule(lr){7-10}
 &  & \multicolumn{1}{c|}{} & \multicolumn{1}{c|}{} & \multicolumn{1}{c|}{} & \multicolumn{1}{c|}{} & \multicolumn{1}{c}{Recall} & \multicolumn{1}{c|}{IoU} & \multicolumn{1}{c}{MAE} & \multicolumn{1}{c|}{RMSE} & \multicolumn{1}{c}{} \\ \midrule

\multirow{8}{*}{\makecell{\\ \\ \\ \\ \\ \\\rotatebox{90}{Task Decomposition}}} & 
\multirow{6}{*}{\makecell{\\ \\ \\ \\ \\ \\ \\ \\ \rotatebox{90}{Static}}} & \begin{tabular}[c]{@{}l@{}}~\cite{Gebrehiwot2022floodDepth_UAV_DEM_FCN}\\ 2021, M\footnotemark[3]\end{tabular} & \begin{tabular}[c]{@{}l@{}}DEM, LiDAR,\\ UAV, Gauge\end{tabular} & 0.026 & \begin{tabular}[c]{@{}l@{}}FCN-8s \\ (CNN)\end{tabular} & - & - & - & 26 & - \\ \cmidrule(l){3-11} 
 &  & 
 \begin{tabular}[c]{@{}l@{}}~\cite{Fang2024DL_MIKE}\\ 2024, M\end{tabular} & \begin{tabular}[c]{@{}l@{}}UAV,\\ MIKE 21\end{tabular} & <0.1 & \begin{tabular}[c]{@{}l@{}}RandLA-NET\\ (CNN)\end{tabular} & - & 70.8 & - & - & - \\ \cmidrule(l){3-11} 
 &  & 
 \begin{tabular}[c]{@{}l@{}}~\cite{Tan2024upscalingDEM_CNN}\\ 2024, M\end{tabular} & \begin{tabular}[c]{@{}l@{}}LR/HR DEM, \\ HR UAV\end{tabular} & \begin{tabular}[c]{@{}l@{}}0.1\\ --\\ 18\end{tabular} & \begin{tabular}[c]{@{}l@{}}ResNet,\\ DeepLabV3\\ (CNN)\end{tabular} & 76.9 & 50.3 & - & - & - \\ \cmidrule(l){3-11} 
 &  & 
 \begin{tabular}[c]{@{}l@{}}~\cite{do2023CGAN}\\ 2023, M\end{tabular} & \begin{tabular}[c]{@{}l@{}}DEM, Rainfall\\ HEC-RAS\end{tabular} & 3 & \begin{tabular}[c]{@{}l@{}}cGAN-Flood \\ (GAN)\end{tabular} & - & 82.5 & - & 10 & \begin{tabular}[c]{@{}l@{}}50--250\end{tabular} \\ \cmidrule(l){3-11} 
 &  & 
 \begin{tabular}[c]{@{}l@{}}~\cite{Hosseiny2020flooddepth_MLP-RF}\\ 2020, S\footnotemark[4]\end{tabular} & \begin{tabular}[c]{@{}l@{}}DEM\\ iRIC\end{tabular} & 5 & \begin{tabular}[c]{@{}l@{}}RF, \\ ANN (MLP)\end{tabular} & - & - & <5 & <22 & 60 \\ \cmidrule(l){3-11} 
 &  & 
 \begin{tabular}[c]{@{}l@{}}~\cite{Yokoya2022DL_modelFusion} *\footnotemark[5]\\ 2022, M\end{tabular} & \begin{tabular}[c]{@{}l@{}}DEM, Satellite, \\ UAV\\ Numerical-model\end{tabular} & 5 & \begin{tabular}[c]{@{}l@{}}Attention \\ U-Net (CNN)\end{tabular} & - & 44.5 & - & 38.5 & - \\ \cmidrule(l){2-11} 
 & \multirow{1}{*}{\makecell{\rotatebox{90}{Dynamic}}} & 
 
 \begin{tabular}[c]{@{}l@{}}~\cite{Zhou2022USRR1DCNN} *\\ 2022, M\end{tabular} & \begin{tabular}[c]{@{}l@{}}DEM\\ TUFLOW\end{tabular} & 10 & \begin{tabular}[c]{@{}l@{}}1D-CNN \\ U-Net (CNN)\end{tabular}  & 97.6 & - & - & 3.4 & 98 \\ \cmidrule(l){3-11} 
 & \multicolumn{1}{l}{} & 
 \begin{tabular}[c]{@{}l@{}}~\cite{Zhou2021DL_SRR_TUFLOW_LSTM} *\\ 2021, M\end{tabular} & \begin{tabular}[c]{@{}l@{}}DEM\\ TUFLOW\end{tabular} & 20 & \begin{tabular}[c]{@{}l@{}}SRR-DL\\  (LSTM)\end{tabular} & 99 & - & - & 80 & 21--86 \\ \midrule

\multirow{13}{*}{\makecell{\\ \\ \\ \\ \\ \\ \\ \\ \\ \\ \\ \\ \rotatebox{90}{End-to-End}}} & 
\multirow{10}{*}{\makecell{\\ \\ \\ \\ \\ \\ \\ \\ \\ \\ \\ \rotatebox{90}{Dynamic}}} & 
\begin{tabular}[c]{@{}l@{}}~\cite{EL2024CNNRealTime}\\ 2024, S\end{tabular} & \begin{tabular}[c]{@{}l@{}}Rainfall\\ Hydro-model\end{tabular} & 2 & CNN & - & - & 0.9 & 3.4 & 120 \\ \cmidrule(l){3-11} 
 &  & 
 \begin{tabular}[c]{@{}l@{}}~\cite{Chaudhary2024AttentionUTAE} *\\ 2024, S\end{tabular} & \begin{tabular}[c]{@{}l@{}}DEM, Rainfall\\ CADDIES\end{tabular} & 2 & \begin{tabular}[c]{@{}l@{}}U-TAE\\ (CNN)\end{tabular} & - & - & 0.75 & - & - \\ \cmidrule(l){3-11} 
 &  & 
 \begin{tabular}[c]{@{}l@{}}~\cite{Shao2024CRU}\\ 2024, M\end{tabular} & \begin{tabular}[c]{@{}l@{}}DEM, Rainfall\\ LISFLOOF-FP\end{tabular} & 3 & \begin{tabular}[c]{@{}l@{}}CRU-Net\\ (CNN)\end{tabular} & 94.2 & 87.1 & 1.1 & 5.4 & - \\ \cmidrule(l){3-11} 
 &  & 
 \begin{tabular}[c]{@{}l@{}}~\cite{Kabir2020CNN_LISFLOOD-FP_Spatial_Tempporal} *\\ 2020, M\end{tabular} & \begin{tabular}[c]{@{}l@{}}DEM (LiDAR)\\ LISFLOOD-FP\end{tabular} & 5 & CNN, SVR & 93 & - & - & 18 & 38 \\ \cmidrule(l){3-11} 
 &  & 
 \begin{tabular}[c]{@{}l@{}}~\cite{Löwe2021UFlood}\\ 2021, S\end{tabular} & \begin{tabular}[c]{@{}l@{}}DEM, Rainfall\\ MIKE 21\end{tabular} & 5 & \begin{tabular}[c]{@{}l@{}}U-FLOOD \\ (CNN)\end{tabular} & - & 58.3 & - & 8 & - \\ \cmidrule(l){3-11} 
 &  & 
 \begin{tabular}[c]{@{}l@{}}~\cite{Fereshtehpour2024CNN_DEM_DTM}\\ 2024, S\end{tabular} & \begin{tabular}[c]{@{}l@{}}LiDAR, Rainfall\\ LISFLOOD-FP\end{tabular} & \begin{tabular}[c]{@{}l@{}}15\\ --\\ 30\end{tabular} & 1D CNN & 99.5 & - & - & 65 & - \\ \cmidrule(l){3-11} 
 &  & 

 \begin{tabular}[c]{@{}l@{}}~\cite{Chu2020ANN-TUFLOW}\\ 2020, S\end{tabular} & \begin{tabular}[c]{@{}l@{}}DEM\\ TUFLOW\end{tabular} & 20 & GRNN & - & - & - & 51 & 100 \\ \cmidrule(l){3-11} 
 &  & 
 \begin{tabular}[c]{@{}l@{}}~\cite{Bentivoglio2023SWE-GNN} *\\ 2023, S\end{tabular} & \begin{tabular}[c]{@{}l@{}}DEM\\ Hydro-model\end{tabular} & 100 & \begin{tabular}[c]{@{}l@{}}SWE-GNN \\ (GNN)\end{tabular} & - & 80+ & <5.8 & <15.4 & 100+ \\ \cmidrule(l){3-11} 
 &  & 
 \begin{tabular}[c]{@{}l@{}}~\cite{Zhou2023LSTMRapidMIKE21}\\ 2023, S\end{tabular} & \begin{tabular}[c]{@{}l@{}}DEM, Rainfall\\ MIKE 21\end{tabular} & - & LSTM & - & - & - & <5 & 19585 \\ \cmidrule(l){2-11} 
 &{\multirow{3}{*}{\makecell{\\ \\ \rotatebox{90}{Static}}}} & 
 
 \begin{tabular}[c]{@{}l@{}}~\cite{Guo2021CNN_CADDIES} *\\ 2021, S\end{tabular} & \begin{tabular}[c]{@{}l@{}}DEM,\\ CADDIES\end{tabular} & 1 & CNN & - & - & <10 & - & 200 \\ \cmidrule(l){3-11}  & \multicolumn{1}{l}{} & 
 
 \begin{tabular}[c]{@{}l@{}}~\cite{Seleem2023CNN_RF_Compare} *\\ 2023, S\end{tabular} & \begin{tabular}[c]{@{}l@{}}DEM\\ TELEMAC\end{tabular} & 1 & \begin{tabular}[c]{@{}l@{}}Unet\\ (CNN)\end{tabular} & - & 60+ & - & <10 & - \\ \cmidrule(l){3-11} & \multicolumn{1}{l}{} & 
 
 \begin{tabular}[c]{@{}l@{}}~\cite{Guo2022catchment} *\\ 2022, S\end{tabular} & \begin{tabular}[c]{@{}l@{}}DEM\\ CADDIES\end{tabular} & 2 & \begin{tabular}[c]{@{}l@{}}Unet \\ (CNN)\end{tabular} & - & - & - & <40 & 6000+ \\ \cmidrule(l){1-11} 

\end{tabular}
      \begin{tablenotes} % footnotes
            \item \noindent\footnotemark[1] Res: Resolution of the input data.
                      \noindent\footnotemark[2] Time: How many times faster than hydrodynamic models.
                      
            \item  \noindent\footnotemark[3] M: Measured data.
                        \noindent\footnotemark[4] S: Simulated data.
                        \noindent\footnotemark[5] *: Open access datasets.  
            \item 
     \end{tablenotes} % footnotes
\end{threeparttable}
\end{table}

%\subsection{Discussion}
As shown in Table \ref{summary}, 3D flood mapping relies heavily on simulated data, compared to 2D mapping, which frequently uses measured data. This reliance is due to the difficulty of obtaining ground truth for flood depth, resulting in a scarcity of publicly available 3D flood datasets based on measured data. Moreover, the lack of standardized datasets poses significant challenges for establishing baselines and comparing the performance of different methods and models.
Despite limitations in datasets, DL models have a significant advantage over traditional models in computational efficiency. They can deliver predictions up to 100 times faster than hydraulic models. This enables real-time 3D flood mapping, which was unachievable due to the time-consuming nature of hydraulic models.

Table \ref{summary} also indicates that within task decomposition methods, studies focus more on static mapping.  The measured data used can have centimeter-level resolutions, and the average RMSE for flood depth prediction is around 20 cm. In contrast, end-to-end methods emphasize dynamic mapping. However, these methods use simulation data with a maximum resolution of 1 m and yield more accurate flood depth predictions, with most studies reporting RMSE below 20 cm. In both methods, CNNs are the most widely used DL models.

\begin{figure}
    \centering
    \includegraphics[width=0.85\linewidth]{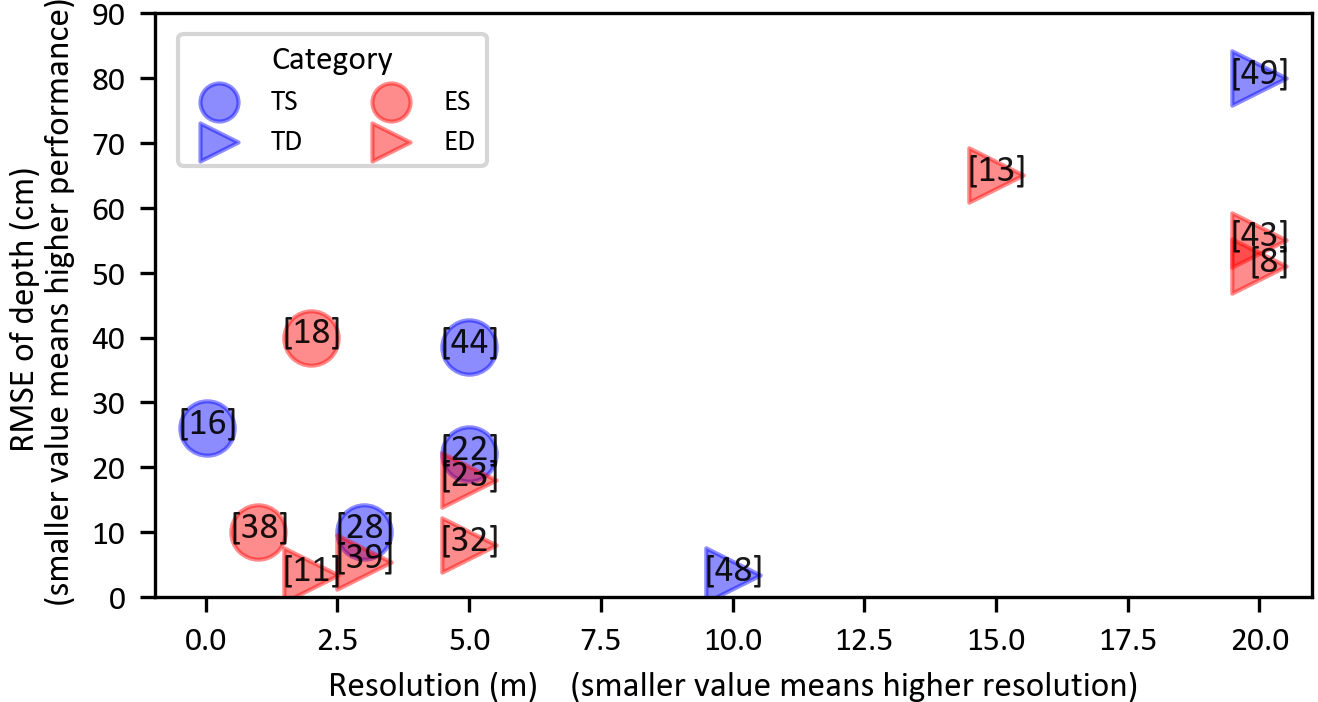}
    \caption{Resolution vs Depth performance. 
    (TS: Task decomposition Static; TD: Task decomposition Dynamic; ES: End-to-end Static; ED: End-to-end Dynamic)}
    \label{fig:metrics}
\end{figure}

Fig.~\ref{fig:metrics} visualizes the depth performance from the literature (input data with resolutions below 20 m per pixel), showing the relationship between RMSE and resolution. The x-axis represents spatial resolution.

As shown in Fig.~\ref{fig:metrics}, there is a clear correlation between RMSE and spatial resolution. Higher resolution generally yields a better RMSE, regardless of whether task decomposition or end-to-end methods are used.
Moreover, the choice of data features also affects RMSE.  At the same resolution, models using dynamic mapping (triangle markers) tend to perform better than those using static mapping (circle markers). This improved performance is likely due to the incorporation of temporal data in dynamic mapping, which offers a more accurate representation of flood characteristics.

\section{Challenges and Future Directions}
This section identifies the major obstacles in advancing DL-based 3D flood mapping and proposes potential solutions.
\subsection{Challenges }
Despite the advancements in 3D flood mapping using DL models, several challenges persist, particularly in data, models, and practical applications.

\paragraph{Challenges in Data.}
The scarcity of 3D datasets is a major obstacle, resulting in a lack of standard baselines for comparing methods. Researchers often have to create their own datasets, which raises the barrier to entry. Some use simulated data from hydraulic models, but this risks training on inaccurate flood characteristics. Moreover, ground truth for flood depth is limited—sparse water gauge records fail to capture comprehensive flood profiles, making model training, validation, and assessment difficult. In addition, satellite imagery often lacks the resolution required for detailed urban mapping, while UAV data, though high-resolution, is constrained by high costs, limited coverage, and logistical issues.

Integrating diverse data sources—such as DEM, multispectral, and hydrological datasets—is challenging due to variations in resolution, coverage, and accuracy. Reliance on simple interpolation methods can lead to data distortion, making effective fusion a persistent issue.

\paragraph{Challenges in DL Models.}

In terms of DL models, while advanced models like GANs and GNNs have emerged, traditional DL models, especially CNNs, still face limitations in robustness, generalization, and interpretability. Hybrid models that combine DL with hydrodynamic approaches show promise by merging computational efficiency with physical accuracy. However, poorly designed integrations may increase complexity without performance gains and can compromise stability. Furthermore, the "black-box" nature of DL models makes it difficult to discern what patterns are being learned. Techniques such as Grad CAM have improved explainability, yet they do not fully alleviate user concerns about model transparency, ultimately affecting trust in these systems.

\paragraph{Challenges in Applications.}
Integrating 3D flood mapping models into practical systems remains underexplored. Current research prioritizes mapping accuracy and speed over real-world deployment. Ethical issues add another challenge because high-resolution spatial data may reveal sensitive information. This raises privacy and consent concerns. Unrepresentative training data can introduce algorithmic bias and lead to flawed flood risk assessments. Consequently, public policies based on these maps must be transparent and accountable, and stakeholders must understand the models' limitations and assumptions.

\subsection{Future Directions}

Despite the challenges, advances in DL and related technologies promise more accurate and efficient models. These improvements are expected to generate higher-resolution maps and support a broader range of applications, from early warning systems to urban and land use planning.

% Addressing Data Scarcity:
Data scarcity remains a significant challenge. Future research should focus on developing concrete solutions to fill this gap. For instance, advanced data augmentation techniques can be developed to synthetically enrich benchmarked datasets. Moreover, establishing collaborative partnerships with governmental agencies and private organizations could facilitate access to more comprehensive, high-resolution data, thereby improving the robustness of flood mapping models.

% \subsubsection{Improved Data Fusion}
Future research should explore advances in multi-data fusion that enable the integration of complementary information from diverse sources such as satellites, DEMs, and weather data. This approach enhances both spatial and temporal resolution while reducing uncertainties in mapping. In particular, fusing multiple data sources improves the accuracy of depth estimation, which is critical for boosting overall performance in 3D flood mapping.

% \subsubsection{Hybrid Models & Interpretability}
Further work is needed to boost model performance. One promising approach is to merge various DL architectures—such as incorporating Vision Foundation Models like the Segment Anything Model (SAM)—to enhance mapping accuracy and usability. Beyond performance, improving DL model interpretability is crucial. Future research should explore advanced techniques that reveal the internal workings of these models, such as sensitivity analysis or integrating physical constraints into DL frameworks. These strategies can enhance both the interpretability and robustness of the results, foster greater user trust, and improve the practical applicability of 3D flood mapping.

% \subsubsection{Policy and Governance implications}
Advances in 3D flood mapping, with its high-resolution and interpretable results, can significantly influence flood management strategies. 
These developments support not only early warning systems and targeted rescue but also urban planning, land use planning, and government flood policies. For instance, they can inform tailored flood insurance policies based on varying risk levels, offering stronger protection for vulnerable regions. While formulating such policies, it is essential to implement robust data governance frameworks, bias mitigation strategies, and transparent stakeholder engagement to address ethical concerns.

\section{Conclusion}
This survey emphasizes the potential of DL in 3D flood mapping, offering significant improvements in computational efficiency and scalability over traditional methods. Task decomposition approaches leverage modular workflows to build on existing 2D research, while end-to-end methods streamline the prediction process. Advanced models, including CNNs, LSTMs, GANs, and GNNs, have demonstrated their ability to address complex flood scenarios, though challenges in datasets, interpretability, model integration, and practical application remain.

Looking ahead, addressing data scarcity is paramount. Innovative solutions like few-shot synthesis and partnering with the government or industry can offer more publicly accessible benchmarked datasets. Researchers should also investigate hybrid models that merge DL with hydrodynamic methods or emerging architectures like SAM. Furthermore, enhancing model transparency through sensitivity analysis and incorporating physical constraints is equally important.

Finally, practical applications of 3D flood mapping must account for ethical and governance challenges. Policy makers should ensure that flood models support transparent and accountable decision-making. Robust data governance and bias mitigation are essential. By bridging current gaps, 3D flood mapping can better support disaster management, urban planning, and policy-making, ultimately reducing flood risks in vulnerable regions.

% \begin{credits}
% \subsubsection{\ackname}  This study was funded

% \subsubsection{\discintname}
% % It is now necessary to declare any competing interests or to specifically
% % state that the authors have no competing interests. Please place the
% % statement with a bold run-in heading in small font size beneath the
% % (optional) acknowledgments\footnote{If EquinOCS, our proceedings submission
% % system, is used, then the disclaimer can be provided directly in the system.},
% % for example: The authors have no competing interests to declare that are
% % relevant to the content of this article. Or: Author A has received research
% % grants from Company W. Author B has received a speaker honorarium from
% % Company X and owns stock in Company Y. Author C is a member of committee Z.
% \end{credits}
\bibliographystyle{splncs04}

\bibliography{refs}





% LNCS Homepage, \url{http://www.springer.com/lncs}, last accessed 2023/10/25
% \end{thebibliography}
\end{document}